\documentclass[10pt,twocolumn,letterpaper]{article}
\usepackage{cvpr}              

\usepackage{graphicx}
\usepackage{amsmath}
\usepackage{amssymb}
\usepackage{booktabs}

\usepackage{threeparttable}
\usepackage{multirow}
\usepackage{amsmath}
\usepackage{caption}

\usepackage{algorithm}
\usepackage{algorithmic}
\usepackage{color}

\definecolor{codeColor}{RGB}{51,153,102}

\usepackage[pagebackref,breaklinks,colorlinks]{hyperref}

\usepackage[capitalize]{cleveref}
\crefname{section}{Sec.}{Secs.}
\Crefname{section}{Section}{Sections}
\Crefname{table}{Table}{Tables}
\crefname{table}{Tab.}{Tabs.}

\def\cvprPaperID{4554} 
\def\confName{CVPR}
\def\confYear{2023}

\begin{document}

\title{Learning Common Rationale to Improve Self-Supervised Representation for Fine-Grained Visual Recognition Problems}


\author{Yangyang Shu\\
\and
Anton van den Hengel\\
\and
Lingqiao Liu\thanks{Corresponding author. This work is supported by the Centre for Augmented Reasoning.}
\and
School of Computer Science, The University of Adelaide \\
{\tt\small \{yangyang.shu,anton.vandenhengel,lingqiao.liu\}@adelaide.edu.au}
}

\maketitle

\begin{abstract}
Self-supervised learning (SSL) strategies have demonstrated remarkable performance in various recognition tasks. However, both our preliminary investigation and recent studies suggest that they may be less effective in learning representations for fine-grained visual recognition (FGVR) since many features helpful for optimizing SSL objectives are not suitable for characterizing the subtle differences in FGVR. To overcome this issue, we propose learning an additional screening mechanism to identify discriminative clues commonly seen across instances and classes, dubbed as common rationales in this paper. Intuitively, common rationales tend to correspond to the discriminative patterns from the key parts of foreground objects. We show that a common rationale detector can be learned by simply exploiting the GradCAM induced from the SSL objective without using any pre-trained object parts or saliency detectors, making it seamlessly to be integrated with the existing SSL process. Specifically, we fit the GradCAM with a branch with limited fitting capacity, which allows the branch to capture the common rationales and discard the less common discriminative patterns. At the test stage, the branch generates a set of spatial weights to selectively aggregate features representing an instance. Extensive experimental results on four visual tasks demonstrate that the proposed method can lead to a significant improvement in different evaluation settings.\footnote{The source code will be publicly available at: \textit{\url{https://github.com/GANPerf/LCR}}}
\end{abstract}

\section{Introduction}
\label{sec:intro}
\begin{figure}[htbp]
	\centering
	\includegraphics[width=3.3in]{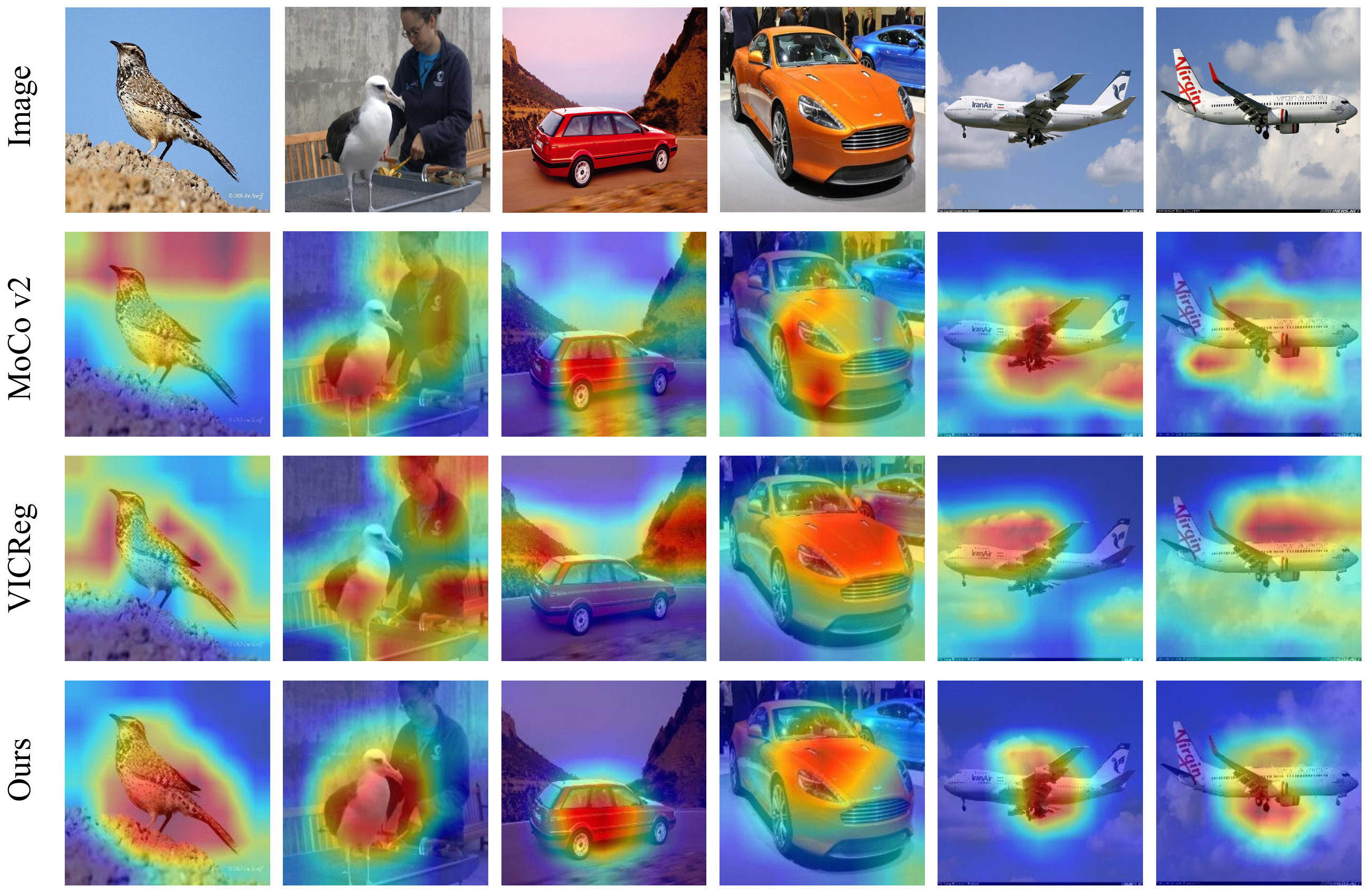}
	\caption{GradCAM~\cite{selvaraju2017grad} visualization for MoCo v2~\cite{he2020momentum}, VICReg~\cite{bardes2021vicreg} and our method on the \texttt{CUB-200-2011}, \texttt{Stanford Cars} and \texttt{FGVC Aircraft} datasets. Compared with the existing method MoCo v2 and VICReg, which are prone to be distracted by background patterns, our method can identify features from the foreground and potentially the key parts of the object.}
	\label{visual_self}
\end{figure}
Recently, self-supervised representations Learning (SSL) has been shown to be effective for transferring the learned representations to different downstream tasks~\cite{bachman2019learning,caron2020unsupervised,he2020momentum,bardes2021vicreg,ericsson2021well}. Methods such as contrastive learning~\cite{chen2020simple,he2020momentum,grill2020bootstrap,chen2021exploring} have demonstrated state-of-the-art feature learning capability and have been intensively studied recently. However, recent studies \cite{cole2022does} suggest that contrastive learning may have a ``coarse-grained bias'' and could be less effective for fine-grained classification problems whose goal is to distinguish visually similar subcategories of objects under the basic-level category. 

This phenomenon is rooted in fine-grained visual recognition (FGVR) properties and the training objective of SSL. SSL tries to minimize a pretext task, e.g., contrastive learning minimizes the distance between same-instance features while maximizing the distance among different-instance features, and any visual patterns that could contribute to loss minimization will be learned. On the other hand, the discriminative patterns for FGVR can be subtle. It is more likely to reside on key object parts. Thus, the feature learned from SSL may not necessarily be useful for FGVR. Figure~\ref{visual_self} shows our investigation of this issue. As seen, the existing SSL paradigms, such as MoCo ~\cite{he2020momentum} and VICReg~\cite{bardes2021vicreg} are prone to learn patterns from irrelevant regions. 
Existing works \cite{zhao2021distilling,selvaraju2021casting,wu2021align} usually handle this issue by recoursing to pre-trained object part detectors or saliency detectors to regularize SSL using patterns from valid regions to achieve the training objective. However, both the part detectors and saliency detectors could restrict the applicability of SSL for FGVR since part detectors are trained from human annotations, and the saliency regions are not always coincident with the discriminative regions.

Therefore, this work aims to directly solve the problem from the target domain data. Specifically, we propose to learn an additional screening mechanism in addition to the contrastive learning process. The purpose of the screening mechanism is to filter out the patterns that might contribute to SSL objectives but are irrelevant to FGVR. Somehow surprisingly, we find that such a screening mechanism can be learned from the GradCAM \cite{selvaraju2017grad} of the SSL loss via an extremely simple method. The whole process can be described as a ``fitting and masking'' procedure: At the training time, we use an additional branch with limited fitting capacity (see more discussion about it in Section \ref{our_method}) to fit the GradCAM calculated from the SSL objective. At the testing time, we apply this additional branch to predict an attention mask to perform weighted average pooling for the final presentation. The motivation for such a design is that the GradCAM fitting branch tends to characterize the discriminative patterns that commonly occur across samples due to its limited fitting capacity, and those common patterns, dubbed common rationale in this paper, are more likely corresponding to the discriminative clues from key object parts or at least foreground regions.

We implement our method based on MoCo v2~\cite{chen2020improved}, one of the state-of-the-art approaches in unsupervised feature learning, which also produces the best performance on FGVR in our setting. Our implementation uses the training objective of MoCo v2 to learn feature representations and derive the GradCAM. Through extensive experiments, we show that our approach can significantly boost the quality of the learned feature. For example, when evaluating the learned feature for retrieval tasks, our method achieves 49.69\% on the \texttt{CUB-200-2011} retrieval task, which is 6.97\% higher than our baseline method (MoCo v2~\cite{chen2020improved}). In the linear evaluation phase, the proposed method achieves new state-of-the-art 71.31\% Top-1 accuracy, which is 3.01\% higher than MoCo v2.

\section{Related Work}
\label{sec:related}
Self-supervised learning aims to learn feature representation from unlabeled data. It relies on a pretext task whose supervision could be derived from the data itself, e.g., image colorization~\cite{zhang2016colorful}, image inpainting~\cite{pathak2016context}, and rotation~\cite{komodakis2018unsupervised} prediction. Contrastive learning is recently identified as a promising framework for self-supervised learning and has been extensively studied \cite{chen2020simple,he2020momentum,wang2020understanding,caron2020unsupervised, dwibedi2021little,ci2022fast}. Despite the subtle differences, most contrastive learning approaches \cite{wang2021dense,zheng2021ressl,gao2022disco} try to minimize the distance between different views of the same images and push away the views of different images. The representative methods are
SimCLR~\cite{chen2020simple} and MoCo~\cite{he2020momentum}. Besides contrastively learning, consistency-based approaches, such as BYOL~\cite{grill2020bootstrap}, SimSiam~\cite{chen2021exploring} and masking-and-prediction-based approaches, such as MAE~\cite{he2022masked}, BEiT~\cite{bao2021beit}, and ADIOS~\cite{shi2022adversarial} are also proven effective for SSL. 

\noindent\textbf{Improving SSL via Better Region Localization.} 
A pipeline to improve the distinguishing ability is to design better data augmentations of SSL. Three such methods were recently proposed: DiLo~\cite{zhao2021distilling}, SAGA~\cite{yeh2022saga}, and ContrastiveCrop~\cite{peng2022crafting}. DiLo uses a copy-and-pasting approach as a kind of augmentation to create an image with a different background. In such a way, the proposed method distills localization via learning invariance against backgrounds and improves the ability of SSL models to localize foreground objects. SAGA adopts self-augmentation with a guided attention strategy to augment input images based on predictive attention. In their method, an attention-guided crop is used to enhance the robustness of feature representation. ContrastiveCrop shows a better crop as an augmentation to generate better views in SSL and keep the semantic information of an image. All of these works locate the object by improving the data augmentations for SSL. In this work, our method is adaptive to locate the key regions for self-supervised learning without needing external augmentations. 
Another family of approaches tries to target the same problem as ours: making the learned feature capture more of the foreground region. CVSA~\cite{wu2021align} proposed a cross-view saliency alignment framework that first crops and swaps saliency regions of images as a novel view generation. Then it adopts a cross-view saliency alignment loss to encourage the model to learn features from foreground objects. CAST~\cite{selvaraju2021casting} encourages Grad-CAM attention to fit the salient image regions detected by a saliency detector. Those methods often rely on pre-trained saliency detection. This implicitly assumes that salient regions are more likely to be discriminative regions. This assumption, however, does not always hold, especially for FGVR. 
\section{Method}
\label{sec:method}
In this section, we first briefly review self-supervised contrastive learning~\cite{chen2020improved} and gradient-weighted class activation mapping (GradCAM)~\cite{selvaraju2017grad} as preliminary knowledge. Then, we introduce the proposed approach.  
\subsection{Preliminary}
\noindent\textbf{Self-Supervised Contrastive Learning.} 
Given an image $I$ from a batch of samples, $x=t(I)$ and $x^{\prime}=t^{\prime}(I)$ are the two augmented views, where the $t$ and $t^{\prime}$ are two different transformations sampled from a set of data augmentations $\cal{T}$. Then the views $x$ and $x^{\prime}$ are used as the input of an encoder $f_\theta$ to generate their feature representations $u=f_\theta(x)$ and $u^{\prime}=f_\theta(x^{\prime})$. Finally, the projector heads $g_\theta$ are used to map $u$ and $u^{\prime}$ onto the embeddings $z=g_\theta(u)$ and $z^{\prime}=g_\theta(u^{\prime})$. Let $(z, z^{\prime})$ be embeddings from the same image and are used as a positive pair. Let  $z_k$ be the embedding from a different image, and $(z, z_k)$ thus composes a negative pair. SimCLR~\cite{chen2020simple} adopts a contrastive loss to maximize the agreement of positive pairs over those of negative pairs. The MoCo~\cite{he2020momentum,abbasi2020compress,ci2022fast} family adopts the same contrastive loss but adds a queue to store the image embeddings to alleviate the memory cost due to the large batch size. Formally, the contrastive loss takes the following form 
 \begin{align}\label{eq:contrastive loss}
     \mathcal{L}_{CL} &= -\log\frac{\exp(z\cdot z^{\prime}/t)}{\sum_{i=0}^Q \exp(z\cdot z_i/t)},
 \end{align}
where $t$ denotes a temperature parameter, and $z_i$ is the embedding in the queue. 

\noindent \textbf{Gradient-weighted 
Class Activation Mapping (GradCAM)} is a commonly used way to produce a visual explanation. It identifies the important image regions contributing to the prediction by using the gradient of the loss function with respect to the feature map or input images. In this paper, we consider the gradient calculated with respect to the last convolutional layer feature maps. Formally, we consider the feature map of the last convolutional layer denoting as $\phi(I) \in \mathbb{R}^{H \times W \times C}$, $H$, $W$ and $C$ are the height, width and number of channels of the feature map, respectively. In standard C-way classification, the GradCAM is calculated by:
\begin{align}
\label{gradcam}
&[\mathrm{Grad\text{-}CAM}(\hat{y})]_{i,j} =  ReLU\left(\mathbf{\alpha}_{\hat{y}}^\top[\phi(I)]_{i,j}\right) \nonumber \\
&where, ~\mathbf{\alpha}_{\hat{y}}=\frac{\partial \mathcal{L}_{CE}(P(y),\hat{y})}{\partial [\phi(I)]_{i,j}}\in \mathbb{R}^C,
\end{align}where $\mathcal{L}_{CE}(P(y),\hat{y})$ is the cross-entropy loss measuring the compatibility between the posterior probability $P(y)$ and ground-truth class label $\hat{y}$\footnote{ $\mathcal{L}_{CE}(P(y),\hat{y}) = \log P(\hat{y})$ for the multi-classification problem and the gradient of $\log P(y)$ is proportional to the gradient of the corresponding logit. Those equivalence forms lead to the different definitions of GradCAM in the literature.}. $[\phi(I)]_{i,j} \in \mathbb{R}^C$ denotes the feature vector located at the $(i,j)$-th grid.  $\mathrm{Grad\text{-}CAM}(\hat{y})$ denotes GradCAM for the $\hat{y}$-th class. $[\mathrm{Grad\text{-}CAM}(\hat{y})]_{i,j}$ refers to the importance value of the $(i,j)$th spatial grid for predicting the $\hat{y}$-th class.

Note that although GradCAM is commonly used for supervised classification problems, it can be readily extended to other problems by changing the corresponding loss function.

\subsection{Our Method: Learning Common Rationale}
\label{our_method}
\begin{figure}[tbp]
	\centering
	\includegraphics[width=3.3in]{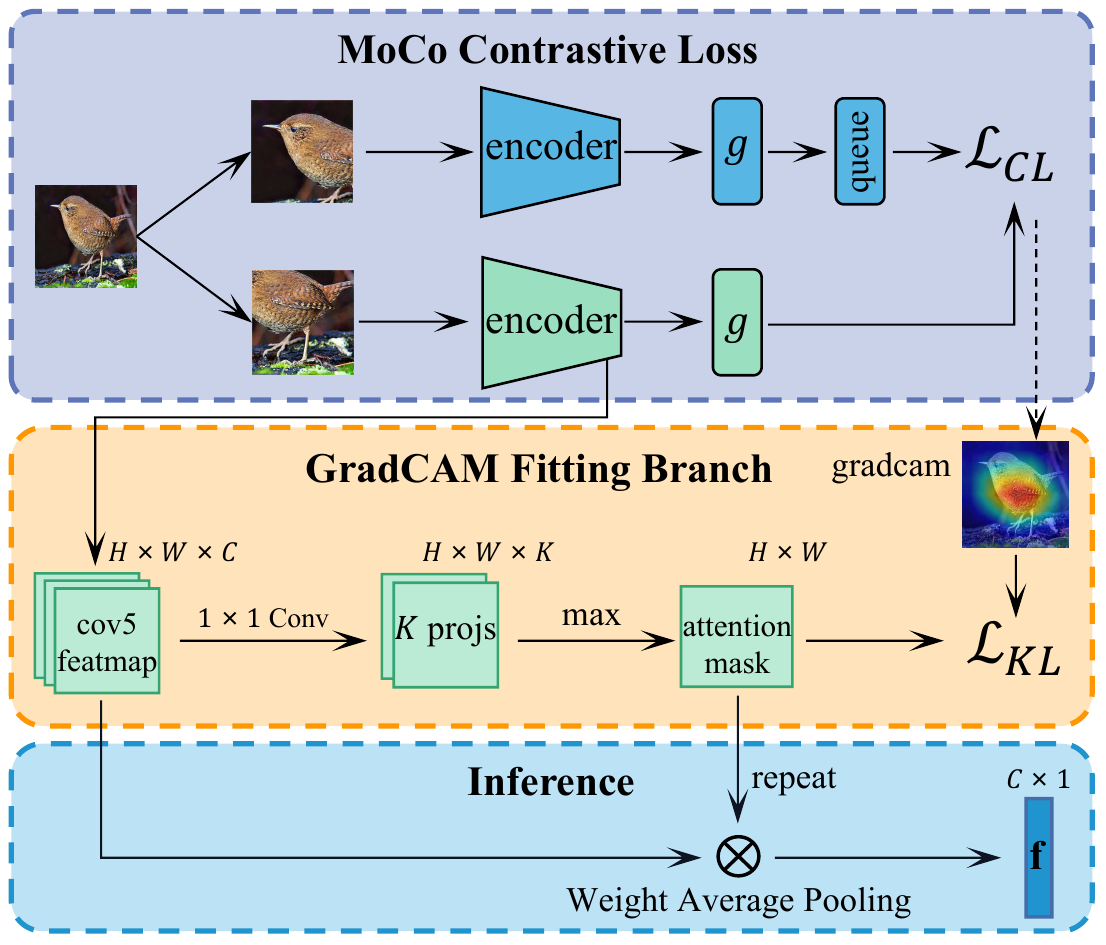}
	\caption{The overview of our method. At the structure level, it has the following components: 1)\textit{Encoder}: MoCo-based contrastive learning is used to create the image-level representation. 2) \textit{GradCAM fitting branch}, which is a convolutional layer with $k$ filters followed by a channel-wise max-out operator. 3) \textit{Inference}: at the inference time, the prediction from GFB is used as a spatial attention mask, and it is used to replace the GAP operation by weighted average pooling to obtain the image representation. }  
	\label{framework}
\end{figure}

Figure \ref{framework} gives an overview of the proposed method. Our method is extremely simple: we merely add one GradCAM fitting branch (GFB) and fit the GradCAM generated from the CL loss at the training time. At the test time, we let this GFB predict (normalized) GradCAM and use the prediction as an attention mask to perform weighted average pooling over the convolutional feature map. The details about this framework are elaborated as follows.

\noindent \textbf{GradCAM Calculation.}
For self-supervised learning, we do not have access to ground-truth class labels. Therefore, we use the contrastive learning loss $\mathcal{L}_{CL}$ in Eq. \ref{eq:contrastive loss} to replace $\mathcal{L}_{CE}$ in Eq. \ref{gradcam}:
\begin{align}
    [\mathbf{G}]_{i,j} = ReLU\left(\frac{\partial \mathcal{L}_{CL}(\psi(\phi(I)))}{\partial [\phi(I)]_{i,j}}^\top[\phi(I)]_{i,j}\right), 
\end{align}where $\psi()$ denotes the feature extractor. 
Note that in the original GradCAM, the GradCAM weight indicates how important each region contributes to classifying the image into the $y$-th class. Similarly, the GradCAM weight from $\mathcal{L}_{CL}$ indicates the contribution of each region to the instance discrimination task in the contrastive learning objective. From Figure \ref{visual_self}, we can see that existing Contrastive Learning approaches learn a diverse set of visual patterns, and not all of them are relevant to the FGVR task.

\noindent \textbf{Architecture of the GFB.}
The structure of the GFB plays an important role in our method. We expect this branch has a limited fitting capability, such that the branch will not overfit the GradCAM but only capture the commonly occurring discriminative patterns, i.e., common rationales. Inspired by \cite{shu2022improving}, we use a convolutional layer with $K$ filter and $1\times1$ kernel size followed by the max-out operation \cite{shu2022improving} as the fitting branch. Formally, such a branch applies the following operation to the local feature $[\phi(I)]_{i,j}$at the $(i,j)$-th grid of the feature map:
\begin{align}
A_{i,j} = \max_k\{ \mathbf{w}^\top_k [\phi(I)]_{i,j}\},
\end{align} where $A_{i,j}$ will be the predicted GradCAM weight at the $(i,j)$-th grid and $\mathbf{w}_k$ is the $k$-th filter (projector) in the convolutional layer. Intuitively, the convolutional layer can be seen as a collection of $K$ detectors and the above operation can be understood as follows: each projection vector $\mathbf{w}_k$ detects an object part $\mathcal{P}_k$; the max-out operation takes the maximum of $K$ projections at a given location, which will result in a high value if \textit{any of the $K$ object parts are detected}. Varying $K$ could adjust the size of the detector pool and influence the fitting capability. 

\noindent \textbf{Training Loss.} In addition to the contrastive learning loss, we require the GFB to produce a similar attention map as the one produced from the GradCAM of $\mathcal{L}_{CL}$. We follow \cite{shu2022improving} to normalize the GradCAM into a probability distribution and adopt KL divergence as the loss function. 

Formally, we normalize the GradCAM $\mathcal{G}$ and $\mathcal{A}$ via the softmax function:
\begin{align}\label{eq: normalization}
    &[\mathbf{\bar{G}}]_{i,j} = \frac{\exp{([\mathbf{G}]_{i,j}/\tau)}}{\sum_{i=1}^H \sum_{j=1}^W \exp{{([\mathbf{G}]_{i,j}/\tau)}}}, \nonumber \\
   & [\mathbf{\bar{A}}]_{i,j} = \frac{\exp{([\mathbf{A}]_{i,j}/\tau)}}{\sum_{i=1}^H \sum_{j=1}^W \exp{{([\mathbf{A}]_{i,j}/\tau)}}},
\end{align}where $[\cdot]_{i,j}$ denotes the $i,j$-th element of the feature map. $\tau$ is an empirical temperature parameter and we set it to 0.4 here. Thus, the objective can be expressed as follows:
 \begin{align}\label{eq:KL}
     \mathcal{L}_{KL} &= \sum_{i=1}^H \sum_{j=1}^W [\mathbf{\bar{A}}]_{i,j} \log\frac{[\mathbf{\bar{A}}]_{i,j}}{[\mathbf{\bar{G}}]_{i,j}},
 \end{align}
Thus, the final loss function taken on all images over an unlabelled dataset is shown as followings.
\begin{align}\label{eq:total}
    \mathcal{L} = \lambda \mathcal{L}_{CL} + \nu  \mathcal{L}_{KL}.
\end{align}

\noindent \textbf{Inference.} At the inference time, the GradCAM fitting branch will be firstly used to produce an attention mask, i.e., prediction of GradCAM. This attention mask is firstly normalized using $A'_{i,j} =  \frac{A_{i,j}-\min (A)}{1e^{-7}+\max(A)}$. Then we use the normalized attention to perform weighted average pooling of features from the last-layer convolutional feature map. Formally, it calculates:
 \begin{align}\label{eq:attention pooling}
    \mathbf{f} = \sum_{i,j} [A']_{i,j} [\phi(I)]_{i,j} \in \mathbb{R}^C.  
 \end{align} $\mathbf{f}$ is used for the downstream tasks.

\noindent \textbf{Discussion.} Our approach is inspired by the SAM method in \cite{shu2022improving}. However, there are several important differences:
\begin{itemize}
\item SAM method was proposed for supervised FGVC under a low-data regime, while our approach is designed for self-supervised learning.
\item Most importantly, our study discovers that for self-supervised feature learning, the best strategy to interact with the GFB and the feature learning branch is different from what has been discovered in \cite{shu2022improving}. Table \ref{comp_SAM_variant} summarizes the main differences. As seen, unlike either SAM or SAM-bilinear, we do not introduce any interaction between the feature learning branch and the GradCAM fitting branch during training but allow them to perform weighted average pooling at the test stage. In fact, we find that applying cross-branch interaction at the training stage will undermine the feature learning process. This is because applying cross-branch interaction, e.g., using $\mathbf{A}$ to weighted pool features will prevent CL from exploring discriminative patterns, especially when $\mathbf{A}$ has not been properly learned. 
\item For multiple projections, SAM uses bilinear pooling to aggregate $\mathbf{A}$ and the original feature map, resulting in high-dimensional feature representation. Our work performs max-out on multiple projections, resulting in a single attention mask for performing weighted average pooling. Consequently, we could achieve a significant reduction of the feature dimensionality.
\end{itemize}

\begin{table*}[htbp]
\caption{Upper part: summary of the major difference between our method and SAM method \cite{shu2022improving}. Lower part: two variants that are also investigated in this work.}
\label{comp_SAM_variant}
	\centering
\begin{tabular}{c|c|cc|c|c}
\hline
\multirow{2}{*}{Method} & \multirow{2}{*}{\begin{tabular}[c]{@{}c@{}}GradCAM\\ Fitting Branch\end{tabular}}& \multicolumn{2}{c|}{Cross-branch Interaction}      & \multirow{2}{*}{\begin{tabular}[c]{@{}c@{}}Feature\\ Dimension\end{tabular}} &\multirow{2}{*}{\begin{tabular}[c]{@{}c@{}}Loss\\ Function\end{tabular}}\\ \cmidrule(l){3-4} 
                 &       & train             & test              &                                                                                \\ \hline\hline
SAM           & 1 proj      &  $\times$            &  $\times$             & $C$               & CrossEntropy                                                                      \\
SAM-Bilinear   & max[$K$ projs]     & bilinear pooling & bilinear pooling & $C$*$K$             & CrossEntropy                                                                        \\

Ours               & max[$K$ projs]    & $\times$                 & weighted average pooling & $C$         & Contrastive                                                                   \\ \hline
\textit{Ours-DualPooling}               & max[$K$ projs]    & weighted average pooling                & weighted average pooling & $C$ & Contrastive  \\ 
\textit{Ours-MultiTask}               & max[$K$ projs]    & $\times$                & $\times$  & $C$ & Contrastive  \\ 
\hline
\end{tabular}
\end{table*}
To make our study comprehensive, we also explore several variants as baseline approaches.
From the comparison with those methods, the benefit of our design could become more evident. For comparison, we use the following architectures as baselines for self-supervised pre-training:

\noindent \textit{SAM-SSL:} this baseline extends SAM by changing its objective function from cross-entropy to the contrastive loss in MoCo V2. It shares the same architecture as SAM~\cite{shu2022improving}, 
where a projection is used as the GFB, and GradCAM fitting is trained as an auxiliary task to contrastive learning.

\noindent \textit{SAM-SSL-Bilinear:} this baseline extends SAM-bilinear by using the contrastive loss in MoCo V2. The cross-branch interaction of \textit{SAM-SSL-Bilinear} follows the original SAM-Bilinear method.

Besides the above two extensions of the SAM methods, we also consider two variants of our method. The first is called \textit{Ours-MultiTask}, which does not perform weighted average pooling at the test stage but merely uses GradCAM fitting as an auxiliary task. Another is called \textit{Ours-DualPooling}, which performs weighted average pooling at both training and testing stages. Those two variants are summarized in Table \ref{comp_SAM_variant}.


\section{Experiments}
\label{sec:experiments}
In this section, we will evaluate our proposed method on three widely used fine-grained visual datasets (\texttt{Caltech-UCSD Birds (CUB-200-2011)}~\cite{wah2011caltech}, \texttt{Stanford Cars}~\cite{krause20133d} and \texttt{FGVC-Aircraft}~\cite{maji2013fine}), and a large-scale fine-grained dataset (\texttt{iNaturalist2019}~\cite{van2018inaturalist}). 
Our experiments aim to understand the effectiveness and the components of the proposed algorithm. 

\subsection{Datasets and Settings}
\textbf{Datasets.} \texttt{CUB-200-2011} contains 11,788 images with 200 bird species, where 5994 images are used for training and 5794 images for testing. \texttt{Stanford Cars} contains 16,185 images with 196 categories, where 8144 images are for training and 8041 images for testing. \texttt{FGVC-Aircraft}  contains 10,000 images with 100 categories, where 6667 images are for training, and 3333 images are for testing. \texttt{iNaturalist2019} in its 2019 version contains 1,010 categories, with a combined training and validation set of 268,243 images. Note that fine-grained visual task focus on distinguishing similar subcategories within a super-category, while there are six super-categories in \texttt{iNaturalist2019}. 

\textbf{Implementation Details.} 
We adopt the ResNet-50~\cite{he2016deep} as the network backbone, which is initialized using ImageNet-trained weights, and build our method on top of MoCo v2~\cite{chen2020improved}. Therefore the SSL loss term is identical to MoCo v2. The momentum value and memory size are set similarly to MoCo v2, i.e., 0.999 and 65536, respectively. The projector head $g_\theta$ in MoCo v2 is composed of two fully-connected layers with ReLU and a third linear layer with batch normalization (BN)~\cite{ioffe2015batch}. The size of all three layers is 2048 $\times$2048$\times$256. We set the mini-batch size as 128 and used an SGD optimizer with a learning rate of 0.03, a momentum of 0.9, and a weight decay of 0.0001. 100 epochs are used to train the feature extractor. The images from the four FGVR datasets are resized to 224$\times$224 pixels during training times. During testing time, images are firstly resized to 256 pixels and then are center cropped to 224$\times$224 on these four FGVR datasets. 

\subsection{Evaluation Protocols}
\label{exp:retrieval}

\begin{figure}[!htbp]
	\centering
	\includegraphics[width=3.3in]{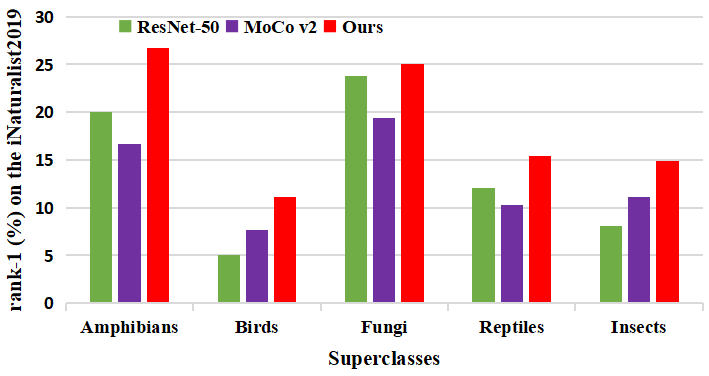}
	\caption{Comparison of ResNet-50, MoCo v2 and Ours on the retrieval rank-1 on the \texttt{iNaturalist2019} dataset. ``\texttt{Amphibians}", ``\texttt{Birds}", ``\texttt{Fungi}", ``\texttt{Reptiles}", and ``\texttt{Insects}" are the superclasses in \texttt{iNaturalist2019}.}
	\label{iNaturalist2019}
\end{figure}

\begin{table*}[htbp]
	\centering
\caption{Classification and retrieval of our method evaluated on the \texttt{CUB-200-2011}, \texttt{Stanford Cars} and \texttt{FGVC Aircraft} datasets. ``ResNet-50'' represents pre-training in the ImageNet dataset~\cite{deng2009imagenet} in a supervised manner, then freezing the ResNet-50 backbone and only optimizing the supervised linear classifier in the classification task. We report the Top 1 and Top 5 (in \%) on the classification task, rank-1, rank-5, and mAP (in \%) on the retrieval task. 100, 50, and 20 are the three different label proportions (in \%) in the classification task.}
\label{classifiaction_and_retrieval}
\begin{tabular}{cc|c|ccc|ccc}
\hline
\multicolumn{2}{c|}{\multirow{2}{*}{Dataset}}                                                                                    & \multirow{2}{*}{Method} & \multicolumn{3}{c|}{Classification} & \multicolumn{3}{c}{Retrieval}                                                                               \\ \cmidrule(l){4-6} \cmidrule(l){7-9}
\multicolumn{2}{c|}{}                                                                                                            &                         & Top 1/Top 5(100)  &Top 1/Top 5(50) &Top 1/Top 5(20)     & rank-1                            & rank-5                            & mAP                               \\ \hline\hline
\multicolumn{2}{c|}{\multirow{3}{*}{\texttt{CUB-200-2011}}}                                                                                      & ResNet-50               & 68.17/90.42  &58.99/85.90&  46.54/77.09            & 48.13                           & 74.12                            & 20.11                            \\
\multicolumn{2}{c|}{}                                                                                                            & MoCo v2                 & 68.30/90.85 &60.96/87.00& 46.91/76.59  & 42.72                           & 69.71                           & 19.68                          \\
\multicolumn{2}{c|}{}                                                                                                            & Ours & \textbf{71.31}/\textbf{92.03} &\textbf{66.52}/\textbf{90.06}&\textbf{55.33}/\textbf{83.52}  & \textbf{49.69} & \textbf{75.23} & \textbf{24.01} \\ \hline
\multicolumn{2}{c|}{\multirow{3}{*}{\texttt{Stanford Cars}}}                                                                                       & ResNet-50               & 57.41/83.55    &46.23/74.31&31.19/58.67           & 27.27                            & 51.90                           & 5.82                            \\
\multicolumn{2}{c|}{}                                                                                                            & MoCo v2                 & 58.43/84.85  &50.17/77.38&35.14/64.10 & 32.22                           & 58.41                           & 8.00                            \\
\multicolumn{2}{c|}{}                                                                                                            & Ours & \textbf{60.75}/\textbf{86.44} &\textbf{53.87}/\textbf{81.72}&\textbf{40.88}/\textbf{69.18}   & \textbf{34.56} & \textbf{60.75} & \textbf{8.87}  \\ \hline
\multicolumn{2}{c|}{\multirow{3}{*}{\texttt{FGVC Aircraft}}}                                                                                   & ResNet-50               & 47.38/74.73   &37.83/67.12&28.20/54.73            & 25.59                            & 46.68                          & 27.31                            \\
\multicolumn{2}{c|}{}                                                                                                            & MoCo v2                 & 52.54/80.74  &45.52/73.85&35.17/65.08 & 32.13                           & 58.71                           & 12.12                            \\
\multicolumn{2}{c|}{}                                                                                                            & Ours& \textbf{55.87}/\textbf{84.73} &\textbf{48.22}/\textbf{77.14}& \textbf{38.55}/\textbf{68.53} & \textbf{34.33} & \textbf{61.09} & \textbf{15.43} \\ \hline
\end{tabular}
\end{table*}
We evaluate the proposed method in two settings: linear probing and image retrieval. Linear probing is a commonly used evaluation protocol in SSL. In linear probing, the feature extractor learned from the SSL algorithm will be fixed and a linear classifier will be trained on top of the learned features. The classification performance of the linear classifier indicates the quality of the learned feature.

Besides linear probing, we also use image retrieval (also equivalent to the nearest neighbor classification task) task to evaluate the learned features, which was also explored in the literature \cite{caron2021emerging,jang2021self,kim2021cds}. This task aims to find the images with the same category as query images based on the learned feature. Note that, unlike linear probing, the image retrieval task does not involve a large amount of labeled data -- which could be used to suppress the less relevant features. In this sense, succeeding in the image retrieval task imposes higher feature quality requirements. Moreover, image retrieval is a practically useful task, and unsupervised feature learning is an attractive solution to image retrieval since the whole retrieval system can be built without any human annotation and intervention. For the retrieval task, we use rank-1, rank-5, and mAP as evaluation metrics. 

\subsection{Main Results}
\label{main_results}
\noindent\textbf{The Effectiveness of the Proposed Method.} Our method is based on MoCo v2. We first compare our result against MoCo v2 to examine the performance gain.
The results are reported in Table~\ref{classifiaction_and_retrieval} and Figure~\ref{iNaturalist2019}. 

From Table~\ref{classifiaction_and_retrieval}, we can see that our method has led to a significant improvement over MoCo v2. It achieves 71.31\% top-1 accuracy and 92.03\% top-5 accuracy on the \texttt{CUB-200-2011} dataset with 100\% label proportions, which is a 3.01\% improvement on top-1 and 1.18\% improvement on top-5 over MoCo v2. Similarly, significant improvement can also be found on the \texttt{Stanford Cars} and \texttt{FGVC Aircraft} datasets. This demonstrates that our methods improve the quality of the feature representations learned from the original MoCo v2. Notably, with the label proportions reduced from 100\% to 20\%, there is generally a better improvement in the performance of the proposed method, which demonstrates that the proposed method learns less noisy features, so that training with fewer data can generalize better.


The advantage of the proposed method can also be seen when evaluating the image retrieval task. As seen from Table~\ref{classifiaction_and_retrieval}, our method leads to a significant boost in performance on all datasets. Specifically, the rank-1, rank-5 and mAP of ours ($K$=32) are 49.69\%, 75.23\%, and 24.01\%, respectively on the \texttt{CUB-200-2011} dataset, which is 6.97\%, 5.52\%, and 4.33\% higher than the method of MoCo v2. Similar improvement can also be observed on the \texttt{Stanford Cars} and \texttt{FGVC Aircraft} datasets. On the large-scale fine-grained dataset, i.e., \texttt{iNaturalist2019}, our method also performs better than MoCo v2 shown in Figure~\ref{iNaturalist2019}. These indicate the proposed method is better for retrieval tasks, which might be due to its ability to filter out less relevant patterns. 

\noindent \textbf{Comparison with Other SSL Frameworks.}
\label{exp:comparision with SSL}
\begin{table*}[htbp]
\caption{Compared to other state-of-the-art self-supervised learning frameworks with Top 1 accuracy on the \texttt{CUB-200-2011}, \texttt{Stanford Cars} and \texttt{FGVC Aircraft} datasets; running time and peak memory on the \texttt{CUB-200-2011} dataset. The training time is measured on 4 Tesla V100 GPUs with 100 epochs, and the peak memory is calculated on a single GPU. Top 1 accuracy (\%) is reported on linear classification with the frozen representations of their feature extractor. For fairness, all following models use the ResNet-50 as the network backbone and initialize the ResNet-50 architecture with ImageNet-trained weights. \textcolor{blue}{blue}=best, \textcolor{green}{green}=second best.}
\label{comp_4SSL}
	\centering
 \begin{threeparttable}
\begin{tabular}{ccccccc}
\hline
Method     &Batch Size& Top 1(\texttt{CUB}) &  Top 1(\texttt{Cars}) & Top 1(\texttt{Aircraft}) & time(\texttt{CUB}) & GPU Memory(\texttt{CUB})\\ \hline
Supervised & - & 81.34      & 91.02       & 87.13  &-&-     \\ \hline
DINO ~\cite{caron2021emerging}    & 32/128& 12.37/16.66             &    9.27/10.51             & 8.52/12.93      &1.5h/1.5h&  4.9G/8.4G     \\
SimCLR~\cite{chen2020simple}\tnote{*}     & 32/128& 33.49/38.39             & 44.31/49.41              &   40.56/45.22      &2.5h/2.5h& 7.1G/23.8G      \\
BYOL~\cite{grill2020bootstrap}\tnote{*}       & 32/128&  36.64/39.27           & 43.66/45.21              &  34.90/37.62      &4.0h/4.0h& 7.4G/24.6G     \\
SimSiam~\cite{chen2021exploring}    & 32/128& 35.82/39.97             &   56.87/{\color{green}58.89}            &  41.59/43.06     &2.0h/2.0h&  4.4G/8.9G     \\
MoCo v2~\cite{chen2020improved}    & 32/128& 68.03/68.30      & 52.61/58.43     & 42.51/{\color{green}52.54}   &2.0h/2.0h&6.1G/10.3G   \\
BarlowTwins~\cite{zbontar2021barlow}    & 32/128&  28.58/33.45     & 23.34/31.91     &  28.35/34.77  &1.5h/1.5h&7.0G/8.9G \\
VICReg~\cite{bardes2021vicreg}    & 32/128&  30.07/37.78     & 19.29/30.80     &  29.97/36.00  &1.0h/1.0h&7.0G/9.0G \\\hline
Ours & 32/128& {\color{green}70.43} /{\color{blue}71.31}     & 56.90/{\color{blue}60.75 }       &  46.92/{\color{blue}55.87 }     &2.0h/2.0h& 6.1G/10.3G \\ 
BYOL+Ours\tnote{*}       & 32/128&  45.79/51.20      &  48.53 /50.64       & 40.08/45.94                &4.0h/4.0h& 7.4G/24.6G     \\\hline

\end{tabular}
\begin{tablenotes}
\footnotesize
    \item[*] Due to computational constraints, we are unable to evaluate on the batch size of 4096 as used in the original paper; we leave this for future work.
\end{tablenotes}
\end{threeparttable}
\end{table*}
In addition to the comparison on MoCo v2, we also compare the proposed method against other commonly used SSL approaches.
We report the Top-1 accuracy, the running time, and the peak memory with two different batch sizes (32 $\&$ 128). The results are shown in Table~\ref{comp_4SSL}. All methods are run on 4 V100 GPUs, each with 32G memory. We report the training speed, GPU memory usage, and performance. For classification performance, the blue marks mean the best classification results, and the green marks mean the second-best classification results. It is clear to see that our method achieves the best Top-1 performance on the \texttt{CUB-200-2011}, \texttt{Stanford Cars}, and \texttt{FGVC Aircraft} datasets. Also, when the GPU resources are limited, e.g., only 1 V100 GPU is available, the proposed method reduces the batch size but still remains a competitive classification performance compared to other SSL methods.
Regarding training speed and GPU memory usage, the proposed method has the same running time as MoCo v2 and SimSiam, slightly more peak memory than SimSiam, but less peak memory, and quicker running time than SimCLR and BYOL. Although DINO uses the least training time and GPU memory usage, their performances are also the worst compared to ours and other self-supervised learning methods. Barlow Twins and VICReg have a quicker running time and less GPU memory than our proposed method with a batch size of 128, but their performances are much worse than ours. 

Our method also can be implemented with other self-supervised learning methods, e.g., BYOL, referring to ``BYOL+Ours'' presented in Table~\ref{comp_4SSL}. As we can see, our method applied to the BYOL objective is consistently superior to the baseline of BYOL on three fine-grained datasets.
\subsection{Comparison with Alternative Solutions}
\label{exp:comparision with AL}

\begin{table*}[htbp]
\caption{The linear Top 1 (\%) and retrieval rank-1 (\%) performance comparisons of recent alternative solutions and Ours on the \texttt{CUB-200-2011}, \texttt{Stanford Cars} and \texttt{FGVC Aircraft} datasets. \textcolor{blue}{blue}=best, \textcolor{green}{green}=second best. Collapse means the model fails to produce meaningful performance.}
\label{comparison_with_crop_table}
\centering
\begin{tabular}{c|c|ccc|ccc}
\hline
\multirow{2}{*}{Method} & Feature   & \multicolumn{3}{c|}{Classification}          & \multicolumn{3}{c}{Retrieval}                 \\ \cmidrule(l){3-5} \cmidrule(l){6-8}
                        & Dimension & \texttt{CUB}      & \texttt{Cars} & \texttt{Aircraft} & \texttt{CUB}  & \texttt{Cars}    & \texttt{Aircraft}  \\ \hline\hline
MoCo v2~\cite{chen2020improved}                 & $C$         & 68.30          & 58.43      & 52.54          & 42.72      & 32.22         & 32.13          \\
MoCo v2 -Bilinear       & $C$*$K$       & 68.44          & 58.06      & 53.01          & 41.27      & 30.89          & 30.80           \\
    \textit{SAM-SSL}                 & $C$         & 68.59          & 58.49      & 52.97          & 42.95      & 32.76          & 32.47          \\
\textit{SAM-SSL-Bilinear}        & $C$*$K$       & \textcolor{green}{71.56} & 59.12      & 55.12          & 44.20      & \textcolor{blue}{35.38} & 32.10           \\
DiLo~\cite{zhao2021distilling,wu2021align}                    & $C$         & 64.14          & -          & -              & -          & -              & -               \\
CVSA~\cite{wu2021align}                   & $C$         & 65.02          & -          & -              & -          & -              & -               \\
LEWEL~\cite{huang2022learning} &$C$*$K$&69.27    &59.02   &54.33 &43.05   & 33.02   &32.88\\
ContrastiveCrop~\cite{peng2022crafting}          & $C$         & 68.82          & \textcolor{green}{61.66}      & 54.40          &43.54            & 32.01               & 32.74                \\\hline
\textit{Ours-DualPooling}                   & $C$         &\multicolumn{3}{c|}{collapse}           & \multicolumn{3}{c}{collapse}   \\
\textit{Ours-MultiTask}             & $C$         & 68.56          &  58.55     &  52.87         &   42.93   & 32.77         & 32.52  \\
Ours                    & $C$         & 71.31          & 60.75      & \textcolor{green}{55.87}          & \textcolor{blue}{49.69}      & \textcolor{green}{34.56}          & \textcolor{green}{34.33}  \\
Ours+ContrastiveCrop    & $C$         & \textcolor{blue}{72.84}          & \textcolor{blue}{63.71}      & \textcolor{blue}{56.08}          &\textcolor{green}{49.36}            &   33.55             & \textcolor{blue}{34.95}                \\ \hline
\end{tabular}
\end{table*}

Our method is featured by its capability of discovering the key discriminative regions, which is vital for FGVR. 
In this section, we compared the proposed method with nine alternative solutions to take object localization into account. The first is to simply use a bilinear network \cite{lin2015bilinear} for MoCo V2. Specifically, we follow the similar bilinear structure as in \cite{shu2022improving}, which implicitly learns $K$ parts and aggregates features from those $K$ parts, and we set $K=32$ to make a fair comparison to us (since we use 32 projections). We still use MoCo v2 as the SSL framework and denote this method MoCo v2 -Bilinear. 
The second and third are the methods of \textit{SAM-SSL} and \textit{SAM-SSL-Bilinear} introduced in Section \ref{our_method} since our method is extended from the self-boosting attention mechanism (SAM) proposed in \cite{shu2022improving}. The fourth and fifth are the two variants considered in Section \ref{our_method}.
The other four comparing methods are the very recently localization-based self-supervised methods, DiLo~\cite{zhao2021distilling}, CVSA~\cite{wu2021align}, LEWEL~\cite{huang2022learning}, and ContrastiveCrop~\cite{peng2022crafting}.
DiLo uses a copy-and-pasting approach as a kind of augmentation to create an image with a different background. The work of CVSA targets a similar problem as ours. They exploit self-supervised fine-grained contrastive learning via cross-view saliency alignment to crops and swaps saliency regions of images. LEWEL adaptively aggregates spatial information of features using spatial aggregation operation between feature map and alignment map to guide the feature learning better. The work of ContrastiveCrop is based on the idea of using attention to guide image cropping, which localizes the object and improves the data augmentation for self-supervised learning.  

The comparison to those nine alternatives is shown in Table~\ref{comparison_with_crop_table}.
As seen, by comparing the \textit{SAM-SSL} and \textit{SAM-SSL-Bilinear}, we observe that the proposed method can lead to overall better performance, achieving a significant boost on some datasets. Occasionally, \textit{SAM-SSL-Bilinear} ($K$=32) can achieve comparable performance as ours, but at the cost of using a much higher feature dimensionality. This clearly shows the advantage of the proposed scheme over the scheme in \cite{shu2022improving}. Furthermore, our method and combined with ContrastiveCrop (i.e., ours+ContrastiveCrop) with the lowest feature dimensions but achieve the best Top 1 and rank-1 performance on the \texttt{CUB-200-2011}, \texttt{Stanford Cars} and \texttt{FGVC Aircraft} datasets, compared to those nine alternatives. 
Also, compared with the other localization-aware SSL methods, our method shows a clear advantage, e.g., ours vs. LEWEL. Finally, we find that the variant of our method does not produce a good performance. For example, when applying weighted average pooling at both training and testing, i.e., \textit{Ours-DualPooling}, will make the feature learning fail completely, and the representations will collapse. On the other hand, not performing cross-branch interaction, i.e., \textit{Ours-MultiTask}, does not bring too much improvement over the MoCo v2 baseline.  

\subsection{The Impact of Number of Projections}

\begin{figure}[htbp]
	\centering
	\includegraphics[width=2.5in]{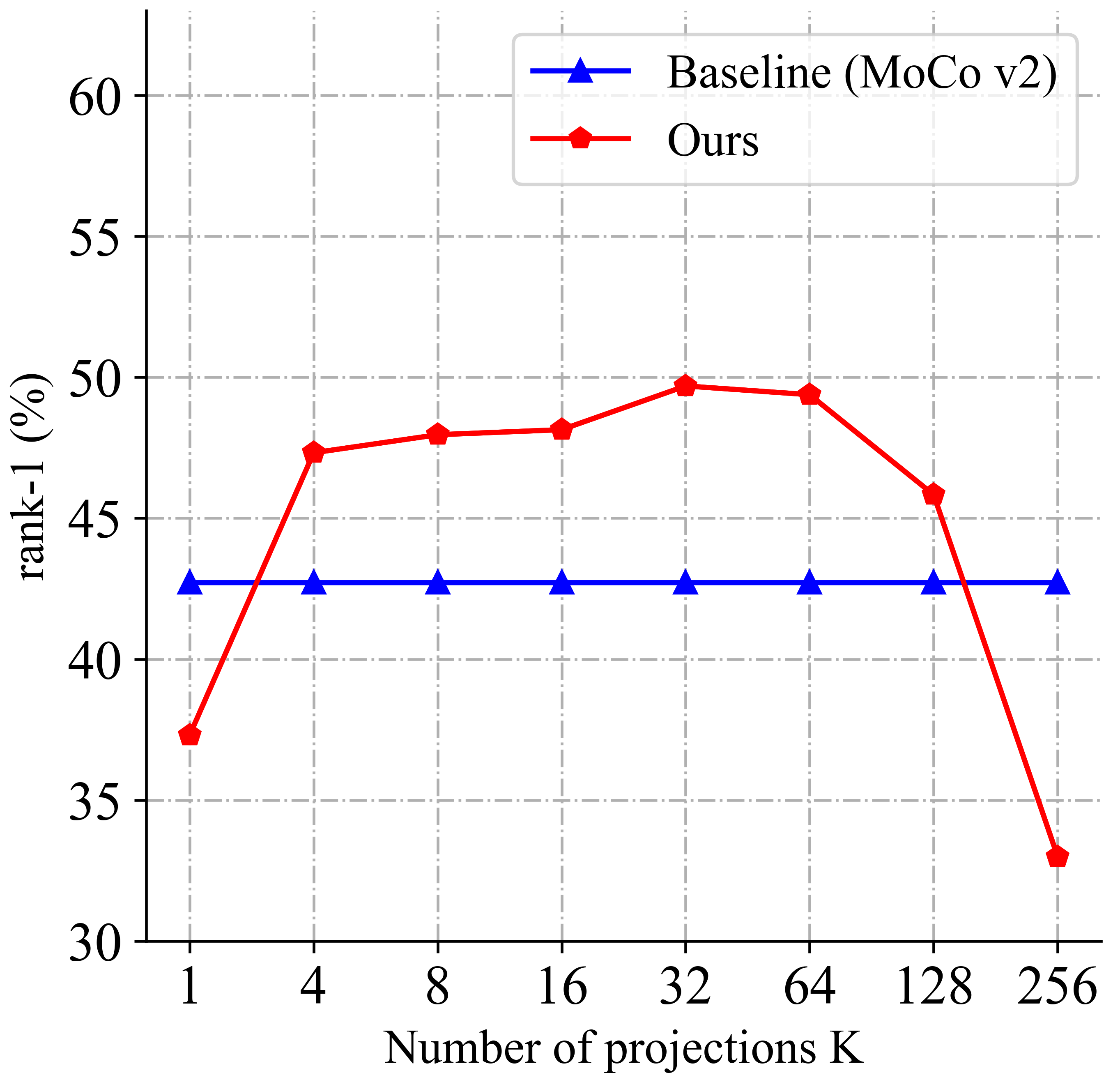}
	\caption{Comparison of MoCo v2 (the blue plot) baseline with our method (the red plot) w.r.t. $K$ on the \texttt{CUB-200-2011}.}
	\label{num_projections}
\end{figure}

To explore the impact of the number of linear projections in our method, we conduct experiments with the different numbers of $K$. Figure~\ref{num_projections} shows the retrieval results of rank-1 w.r.t. eight different projections on the fine-grained dataset \texttt{CUB-200-2011}. As we can see, with the increase of linear projections $K$, the rank-1 gradually increases—the final rank-1 peaks at 49.69\% with $K$ around 32. After $K$ reaches 64, the performance decreases. When the number of projections is very large, the combination of $K$-part detectors becomes spurious and overfits the correlated pattern, thus resulting in a drop in performance. From the curve, we can also see that choosing any value between 4 and 64 can lead to similar performance. So our method is not very sensitive to $K$ once it falls into a reasonable range.  

\subsection{Alternative Structure for GFB}
\label{sec:com_with_mlp}
\begin{table}[htbp]
\caption{Retrieval performance (\%) of our methods and using MLP as the alternative GFB branch. The evaluation is on  the \texttt{Stanford Cars} dataset.}
\label{mlp}
	\centering
\begin{tabular}{ccccc}
\hline
Dataset                   & Architecture    & rank-1 & rank-5 & mAP  \\ \hline\hline
\multirow{2}{*}{\texttt{Cars}}      & Ours            & 34.56  & 60.75  & 8.87 \\
                          & MLP  & 25.57  & 50.91  & 5.92 \\ \hline
\end{tabular}
\end{table}
In this section, we investigate alternative designs for GFB. In particular, we consider using 
a two-layer (with 32 as the intermediate feature dimension) multi-layer perception (MLP) to replace the maximized projections in the proposed method. Compared with our GFB structure, an MLP has better fitting capacity due to the extra linear layer. We conduct experiments on the \texttt{Stanford Cars} dataset, and the result is shown in Table~\ref{mlp}. We find that the performance dramatically decreases when using MLP. This demonstrates the importance of our GFB module design.


\section{Conclusion, Further Results (Appendix) , Limitation and Future Work}
\label{sec:conclusion}
In this paper, we introduce a simple-but-effective way of learning an additional screening mechanism to self-supervised feature learning for fine-grained visual recognition. The idea is to identify discriminative clues commonly seen across instances and classes by fitting the GradCAM of the SSL loss with a fitting-capability-limited branch. Our method achieves state-of-the-art in the classification task and retrieval tasks on fine-grained visual recognition problems. 
More experimental results, including adopting our method for non-fine-grained visual recognition problems, and visualizing each projection in the GFB, can be found in the Appendix. So far, the proposed method seems to be most effective for FGVR, which could be a limitation and we plan to extend the applicability of the proposed method in our future work.



{\small
\bibliographystyle{ieee_fullname}
\bibliography{egbib}
}

\newpage
\appendix
\section{Algorithm} 
\setcounter{algorithm}{0}  
\renewcommand\thealgorithm{\Alph{section}\arabic{algorithm}} 
\begin{algorithm*}[htbp]
	\caption{Pseudocode of Learning Common Rationale on MoCo, PyTorch-style.}
	\label{alg_l1}
	\begin{algorithmic}
		\STATE \textcolor{codeColor}{$\#$ $\psi_q$, $\psi_k$: encoder networks for query and key} \\
        \STATE \textcolor{codeColor}{$\#$ queue: dictionary as a queue of $Q$ keys ($C\times Q$)}\\
        \textcolor{codeColor}{$\#$ $m$: momentum}\\
        \textcolor{codeColor}{$\#$ $t$: temperature in contrastive loss}\\ 
        \textcolor{codeColor}{$\#$ $\tau$: temperature in KL Divergence loss}\\ 
        \STATE \text{ } \\
        \STATE \textcolor{codeColor}{$\#$ Initialize key network parameters} \\
		\STATE $\psi_k$.params = $\psi_q$.params
        \STATE \textcolor{codeColor}{$\#$ Load a minibatch $x$ with $N$ samples} \\
			\FOR {$x$ in dataloader: }
			{
                \STATE \textcolor{codeColor}{$\#$ two different random augmentations} \\
				\STATE $x_q$ = aug($x$) \\
                \STATE $x_k$ = aug($x$) \\
                \STATE \textcolor{codeColor}{$\#$**************************************} \\
                \STATE \textcolor{codeColor}{$\#$ calculate MoCo contrastive loss} \\
                \STATE \textcolor{codeColor}{$\#$**************************************} \\
                \STATE $q$ = $\psi_q$.forward($x_q$) \textcolor{codeColor}{$\#$ queries: $N\times C$}\\       
                \STATE $k$ = $\psi_k$.forward($x_k$)\textcolor{codeColor}{$\#$ keys: $N\times C$} \\
                \STATE $k$ = $k$.detach()\textcolor{codeColor}{$\#$ no gradient to keys} \\
                \STATE \textcolor{codeColor}{$\#$ positive and negative logits.} \\
                \STATE logits$\_$pos = bmm($q$.view($N$,1,$C$), $k$.view(N,C,1))\textcolor{codeColor}{$\#$ shape:$N\times 1$} \\
                \STATE logits$\_$neg = mm($q$.view($N$,$C$), queue.view(C,Q))\textcolor{codeColor}{$\#$ shape:$N\times Q$} \\
                \STATE logits = cat([logits$\_$pos, logits$\_$neg], dim=1) \textcolor{codeColor}{$\#$ shape:$N\times (1+Q)$}\\
                \STATE \textcolor{codeColor}{$\#$ MoCo contrastive loss, positives are the 0-th.} \\
                \STATE labels = zeros($N$) \\
                \STATE loss = CrossEntropyLoss(logits/$t$, labels) \\
                \STATE
                \STATE \textcolor{codeColor}{$\#$**************************************} \\
                \STATE \textcolor{codeColor}{$\#$calculate KL loss} \\
                \STATE \textcolor{codeColor}{$\#$**************************************} \\
                \STATE \textcolor{codeColor}{$\#$get feature map in query network.} \\
                \STATE feat$\_$map=get$\_$cov5($\psi_q$.params) \textcolor{codeColor}{$\#$shape:$N\times 7 \times 7 \times 2048$}\\
                \STATE \textcolor{codeColor}{$\#$calculate gradient of feature map w.r.t. logits$\_$pos.} \\
                \STATE feat$\_$grads=autograd.grad(logits$\_$pos,feat$\_$map)\textcolor{codeColor}{$\#$shape:$N\times 7 \times 7 \times 2048$}\\
                \STATE \textcolor{codeColor}{$\#$calculate GradCAM map} \\
                \STATE gradcam$\_$map=get$\_$gradcam(feat$\_$grads) \textcolor{codeColor}{$\#$shape:$N\times 7\times7$}\\
                \STATE \textcolor{codeColor}{$\#$calculate attention mask} \\
                \STATE attention$\_$mask=projections$\_$max(feat$\_$map)\textcolor{codeColor}{$\#$shape:$N\times 7\times7$} \\
                \STATE \textcolor{codeColor}{$\#$KL loss: attention$\_$mask and gradcam$\_$map} \\KL$\_$loss=kl$\_$div((attention$\_$mask.view($N$,-1)/$\tau$).softmax(dim=-1).log(),(gradcam$\_$map.view($N$,-1)/$\tau$).softmax(dim=-1)) \\
                \STATE loss+=$\nu$*KL$\_$loss \\
                \STATE loss.backward() \\
                \STATE
                \STATE \textcolor{codeColor}{$\#$SGD update for query network} 
				\STATE update ($\psi_q$.params)\\
                \STATE \textcolor{codeColor}{$\#$moment update for key network} \\
                \STATE $\psi_k$.params = $m$*$\psi_k$.params + (1-$m$)*$\psi_q$.params\\
                \STATE \textcolor{codeColor}{$\#$update dictionary: enqueue and dequeue}\\
                \STATE enqueue(queue, $k$)\\
                \STATE dequeue(queue)\\
			}
			\ENDFOR
	\end{algorithmic}
\end{algorithm*}
We present the pseudocode of our method in algorithm~\ref{alg_l1}.

\section{Understanding the Role of Each Projection}

\setcounter{figure}{0}  
\renewcommand\thefigure{\Alph{section}\arabic{figure}} 
\begin{figure}[htbp]
	\centering
	\includegraphics[width=3.4in]{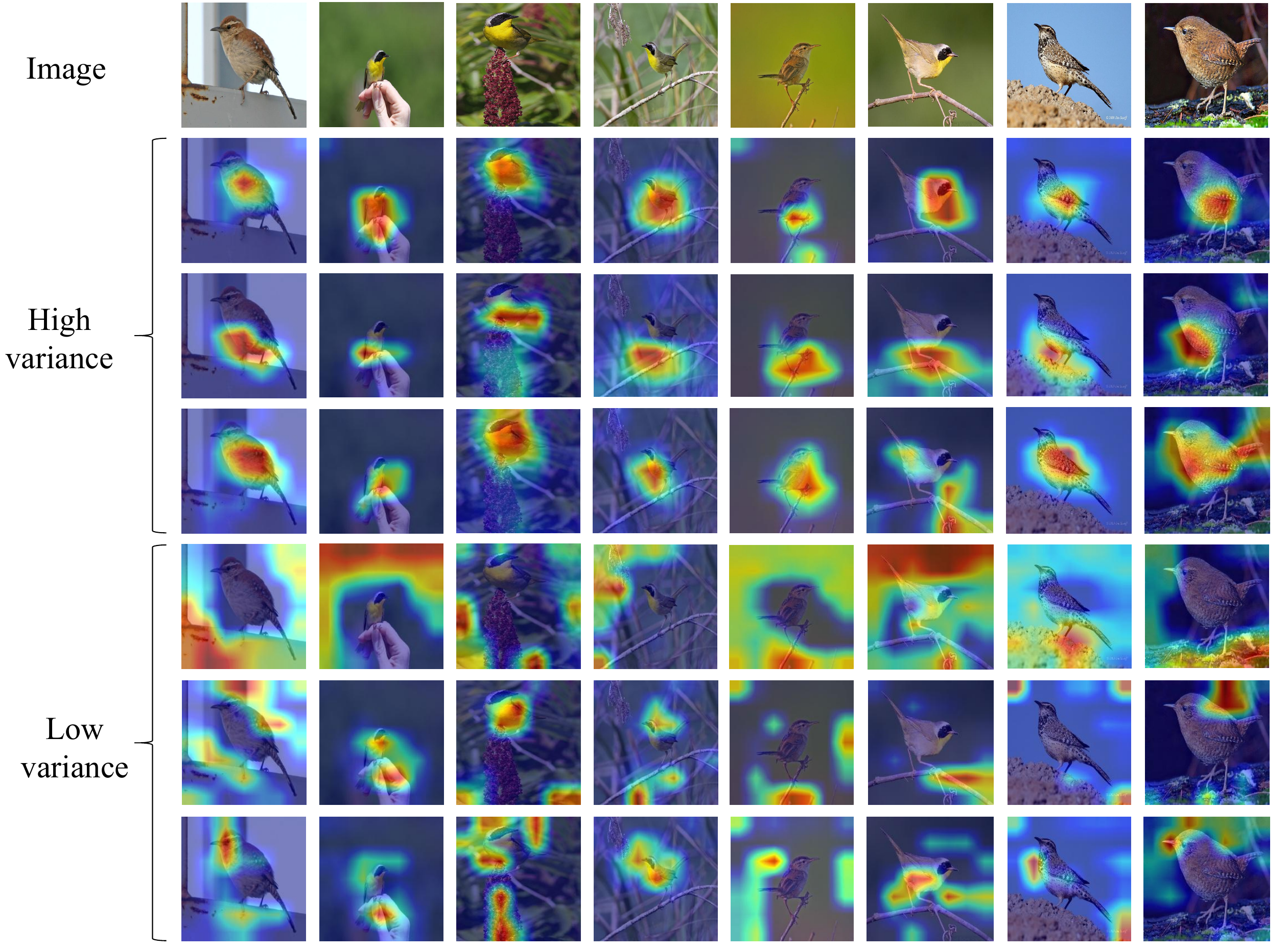}
	\caption{Visualization of high-and-low-variance projections of our method. The first row is the original input images. The 2-4 rows are the visualization of the top three high-variance linear projections, and the 5-7 rows are the visualization of the three projections corresponding to the lowest variance. Note that the colour of the heat map is normalized. It only indicates the relative strength within images, not the absolute value. }
	\label{variance}
\end{figure}

\begin{figure}[htbp]
	\centering
	\includegraphics[width=3.4in]{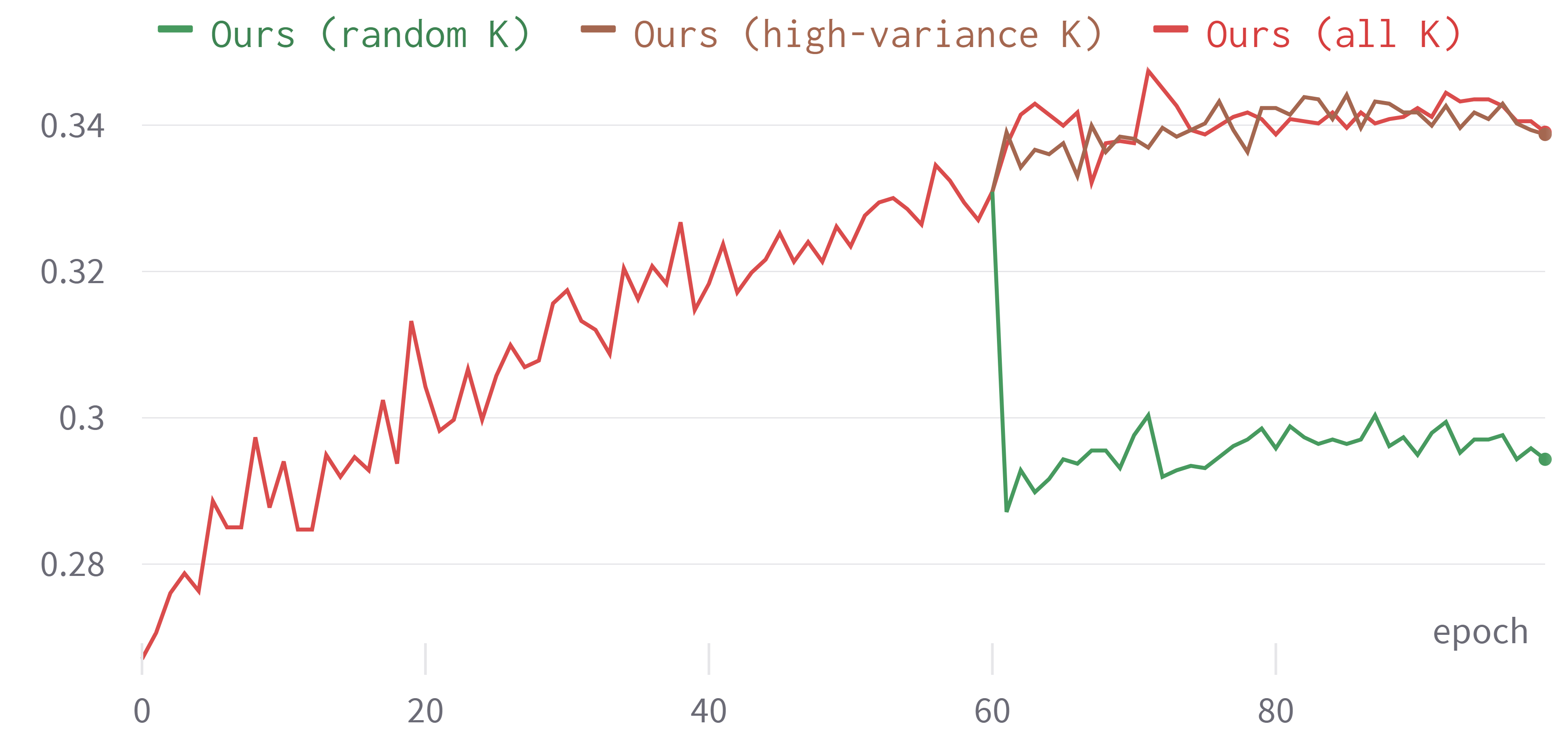}
	\caption{The rank-1 of retrieval task of our methods with different projections selection on the \texttt{FGVC Aircraft} datasets. ``high-variance $K$" means selecting the first eight high-variance projections from 32 projections; ``random $K$" represents randomly selecting eight projections from the remaining 24 projections after exclusive the first eight high-variance projections; ``all $K$" represents all 32 projections selected. }
	\label{rationale}
\end{figure}

In this section, we explore the role of the $K$ linear projections in the GradCAM Fitting Branch (GFB). To better analyze the importance of each projection, we consider the variance of the output of each projection. Higher variance indicates strong signal strength. Figure~\ref{variance} visualize the output of the projection with the three highest variances and three lowest variances. It can be seen that projections with higher variance tend to produce more meaningful output. For example, the projection with the highest variance seems to capture clues that humans rely on to judge fine-grained objects, e.g., the head, body, and back of birds. In contrast, projections with the lowest variance barely attend the reasonable regions. 

Figure~\ref{rationale} shows another experiment to verify the role of different projections. In Figure~\ref{rationale}, the curve shows the rank-1 accuracy of retrieval task on 100 epochs on the \texttt{FGVC Aircraft} datasets. From 0 to 60 epochs, we use all 32 projections ($K$=32) and find the rank-1 accuracy increases significantly. After 60 epochs, we test the three different settings: our method still with all 32 projections, referring as ``Ours (all $K$)" in the red line; our method with the first eight high-variance projections, referring as ``Ours (high-variance $K$)" in brown line; our method with random eight projections chosen from the remaining 24 projections after discarding the top-eight high-variance projections, referring as ``Ours (random $K$)" in the green line. As we can see, the rank-1 greatly increase then gradually becomes stable in the method of Ours (all $K$) and Ours (high-variance $K$), while the rank-1 drop sharply then keep stable in the method of Ours (random $K$). This indicates that the top variance projection preserves the most discriminative rationales. Interestingly, the role of those top-variance projections is akin to that of the top eigenvectors in Principal Component Analysis.

\section{Performance on Non-fine-grained Dataset}
\setcounter{table}{0}
\renewcommand\thetable{\Alph{section}\arabic{table}} 
\begin{table}[htbp]
\caption{Retrieval performance (\%) of our proposed method vs. MoCo v2 on non-fine-grained dataset (\texttt{CIFAR-100}).}
\label{exp_cifar}
	\centering
\begin{tabular}{c|ccc}
\hline
\multirow{2}{*}{Method} & \multicolumn{3}{c}{Metrics} \\ \cline{2-4} 
                      & rank-1   & rank-5  & mAP    \\ \hline
MoCo v2                      & 24.78    & 48.20  & 6.25   \\
Ours                 & 27.14    & 51.25   & 6.44   \\ \hline
\end{tabular}
\end{table}

To observe the performance of our proposed method on the non-fine-grained dataset, we conduct experiments of our method vs. MoCo v2 on \texttt{CIFAR-100} dataset shown in Table~\ref{exp_cifar}. From Table~\ref{exp_cifar}, we can see that our method also works for non-fine-grained cases.


\section{The Impact of Number of Projections on Non-fine-grained Dataset.}
\setcounter{figure}{0}  
\begin{figure}[htbp]
	\centering
	\includegraphics[width=3.0in]{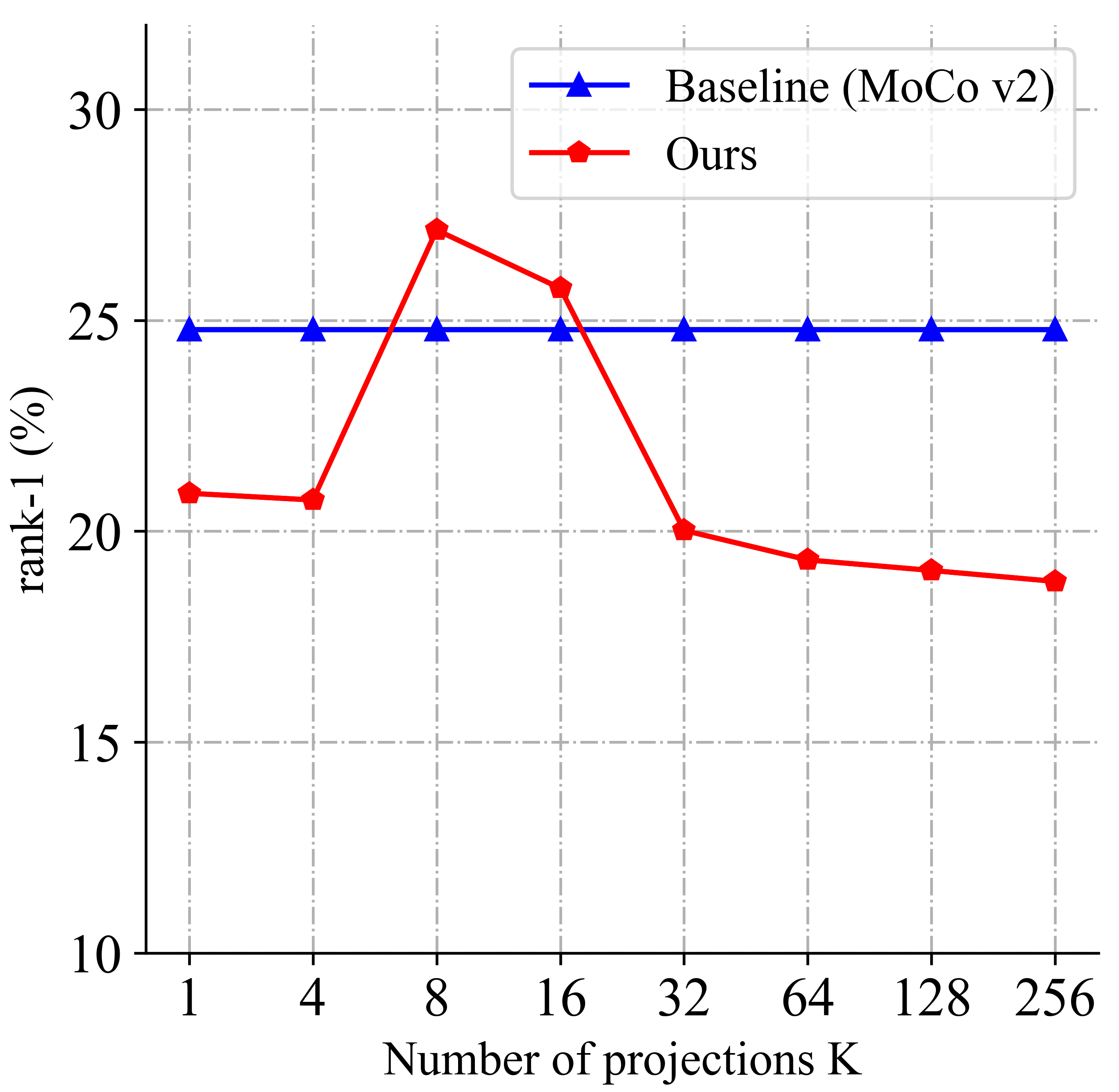}
	\caption{Comparison of MoCo v2 (the blue plot) baseline with our method (the red plot) w.r.t. $K$ on the \texttt{CIFAR-100} dataset.}
	\label{num_projections}
\end{figure}
To explore the impact of the number of linear projections on the non-fine-grained dataset \texttt{CIFAR-100}, we conduct experiments with the different numbers of $K$. Figure~\ref{num_projections} shows the retrieval results of rank-1 w.r.t. eight different projections. As we can see, the rank-1 peaks at 27.14\% with $K$ around 8. With the increase of $K$ from 32, the performance decreases and then gradually becomes stable.

\section{The Impact of GFB Structure on Non-fine-grained Dataset.}
\setcounter{table}{0}
\renewcommand\thetable{\Alph{section}\arabic{table}} 
\begin{table}[htbp]
\caption{Retrieval performance (\%) of our methods and using MLP as the alternative GFB branch. The evaluation is on the \texttt{CIFAR-100} dataset.}
\label{mlp}
	\centering
\begin{tabular}{ccccc}
\hline
Dataset                   & Architecture    & rank-1 & rank-5 & mAP  \\ \hline
\multirow{3}{*}{\texttt{CIFAR-100}} & Ours($K$=8)            & \textbf{27.14}  & \textbf{51.25}  & \textbf{6.44} \\
& Ours($K$=32)            & 20.02  & 40.52  & 4.35 \\
                          & MLP  &19.71   &40.32   & 4.34 \\\hline
\end{tabular}
\end{table}
In this section, we investigate the impact
of GFB structure for the non-fine-grained case by evaluating on \texttt{CIFAR-100}. We follow a similar experimental approach as in the main paper by varying the number of projections. As seen from Figure \ref{num_projections}, the best number of projections seems to be 8, but further increasing the number of projections to 256 does not lead to a significant drop as the case for fine-grained dataset (refer to Figure 4 in the main paper). 

Also, we consider whether we can use multi-layer perception (MLP) to replace the maximized projections in the proposed method shown in Table~\ref{mlp}. We find MLP performs similarly to our methods when $K \geq 32$. This is also different from the case in the fine-grained case. That observation indicates that GFB might have a different characteristic for non-fine-grained data. We plan to leave this in our future work.

\section{Ablation study of weights $\lambda$ and $\nu$.}

\setcounter{table}{0}
\renewcommand\thetable{\Alph{section}\arabic{table}} 
\begin{table}[htbp]
\caption{Retrieval performance (\%) w.r.t the impact of weights $\lambda$ and $\nu$ on the \texttt{CUB-200-2011} dataset.}
\label{regularization}
	\centering
\begin{tabular}{lccc}
\hline
Method                              &$\lambda$    &  $\nu$     & rank-1   \\ \hline
\multirow{4}{*}{$\mathcal{L}_{CL}$+$\mathcal{L}_{KL}$ (Ours)} & 1  & 1     & 47.52    \\
                                    & 1  & 0.1   & 48.83    \\
                                    & 1  & 0.01  & \textbf{49.69}    \\
                                    & 1  & 0.001 & 46.63    \\ \hline
\end{tabular}
\end{table}
Table~\ref{regularization} reports the retrieval performance for different values of the loss term coefficients in the proposed method, where $\lambda$ and $\nu$ control the weight of $\mathcal{L}_{CL}$ and $\mathcal{L}_{KL}$, respectively. We set the weight of $\mathcal{L}_{CL}$ as 1 using a simple grid search method to find that the weight of $\mathcal{L}_{KL}$ as 0.01 is the best. 
\section{The Impact of Embedding Dimension.}

\setcounter{table}{0}
\renewcommand\thetable{\Alph{section}\arabic{table}} 
\begin{table}[htbp]
\caption{Rank-1 accuracy (\%) of our method with different embedding dimensionalities on the retrieval task with 100 pretraining epochs on the \texttt{CUB-200-2011} dataset.}
\label{dimension}
	\centering
\begin{tabular}{cccccc}
\hline
Dimensionality &64 & 128   & 256   & 512   & 1024    \\ \hline
rank-1    & 47.48 & 48.72 & \textbf{49.69} & 49.46 &48.55  \\ \hline
\end{tabular}
\end{table}
Our method is based on the network structure of MoCo v2, and thus we observe the dependency on the dimensionality of the embedding vector shown in Table~\ref{dimension}. Compared to MoCo v2 using 128 as the dimensionality, the impact of embedding vector in our method is different. The performance on the \texttt{CUB-200-2011} dataset increases firstly from $47.48\%$ rank-1 to $49.69$ rank-1 on retrieval task with the dimensionality from 64 to 256. After 256, the performance decreased. Thus, our method uses 256 as the dimensionality of the embedding vector.

\end{document}